\documentclass[runningheads]{llncs}

\usepackage{amsmath,amssymb}
\usepackage{caption}
\usepackage{subcaption}
\usepackage{graphicx}
\usepackage{cite}
\usepackage{hyperref}

\usepackage[capitalize]{cleveref}
\crefname{section}{Sec.}{Secs.}
\Crefname{section}{Section}{Sections}
\Crefname{table}{Table}{Tables}
\crefname{table}{Tab.}{Tabs.}
\crefname{figure}{Figure}{Figures}
\crefname{figure}{Fig.}{Figs.}

\begin{document}
\title{Semantic segmentation with coarse annotations}
%
%
\author{Jort de Jong \and Mike Holenderski}
\authorrunning{Jort de Jong et al.}
%
\institute{Eindhoven Univeristy of Technology, Eindhoven, The Netherlands\\
\email{m.holenderski@tue.nl}}
\maketitle              
\begin{abstract}
Semantic segmentation is the task of classifying each pixel in an image. Training a segmentation model achieves best results using annotated images, where each pixel is annotated with the corresponding class. When obtaining fine annotations is difficult or expensive, it may be possible to acquire coarse annotations, e.g. by roughly annotating pixels in an images leaving some pixels around the boundaries between classes unlabeled. Segmentation with coarse annotations is difficult, in particular when the objective is to optimize the alignment of boundaries between classes. This paper proposes a regularization method for models with an encoder-decoder architecture with superpixel based upsampling. It encourages the segmented pixels in the decoded image to be SLIC-superpixels, which are based on pixel color and position, independent of the segmentation annotation. The method is applied to FCN-16 fully convolutional network architecture and evaluated on the SUIM, Cityscapes, and PanNuke data sets. It is shown that the boundary recall improves significantly compared to state-of-the-art models when trained on coarse annotations.

\keywords{semantic segmentation \and boundary alignment \and coarse annotations \and superpixels}
\end{abstract}

\section{Introduction}\label{intro}

\begin{figure}[htbp]
    \centering
    \subfloat[\centering Fine annotation]{{\includegraphics[width=0.45\textwidth]{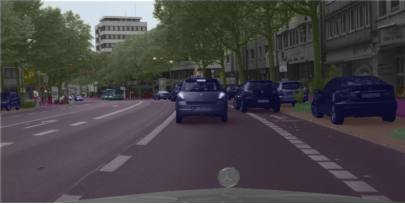}}}%
    \qquad
    \subfloat[\centering Coarse annotation]{{\includegraphics[width=0.45\textwidth]{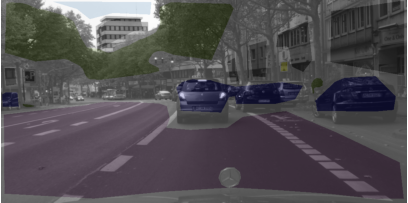}}}%
    \caption{A fine and a coarse annotation from the Cityscapes dataset}%
    \label{label_sample}%
\end{figure}

Semantic segmentation is the task of classifying each pixel in an image. Most literature on semantic segmentation focuses on \emph{fine annotations}, where each pixel in the image is labeled. Collecting fine annotations requires significant effort. Consequently, many data collection approaches prefer \emph{coarse annotations} over fine annotations. Coarse annotations could include object-level annotations (e.g. bounding boxes) or image-level annotations (e.g. image categories). This work considers coarse annotations, which can be generated by roughly annotating images with simple polygons, assigning a common class to every pixel within a polygon. Pixels not included in any of the drawn polygons are designated as unlabeled pixels. Typically, unlabeled pixels tend to be located in the vicinity of class boundaries, as illustrated by \cref{label_sample}.

Models trained on coarse annotations often suffer from imprecise segmentation, where the predicted class boundaries do not align with the ground truth class boundaries. This is referred to as poor boundary alignment. Precise boundary alignment is important for various applications, including semantic segmentation of medical images (e.g. classification of cancerous cells in histopathology), quality control in high-tech manufacturing (e.g. measuring the alignment of assembled components with respect to the reference design), or photo/video editing (e.g. blurring sensitive areas in an image). However, boundary alignment has received relatively little attention in the literature, which mainly focuses on optimizing segmentation accuracy over the entire image.

Semantic segmentation is dominated by deep neural network models. In particular, encoder-decoder architectures, such as Fully Convolutional Networks (FCN) \cite{FCN}, have been shown to be effective on this task. In these models, the encoder extracts high level features from the image into a low resolution encoding, and the decoder is tasked with upsampling the encoding into a high resolution image segmentation. Many popular segmentation models employ this encoder-decoder structure \cite{FCN, FPN, SegNet, UNet, Deconvolution}. Recently, \cite{Imp_integration} proposed to substitute the decoder for superpixel based upsampling. A superpixel is a local cluster of pixels. Superpixels are a useful representation for reducing the complexity of image data and are common in many computer vision tasks. In \cite{Imp_integration}, superpixels are learned based on the encoder features. Incorporating superpixels into FCN-16 results in the HCFCN-16 architecture. The low resolution encoder prediction is upsampled with the superpixels. Substituting the decoder for superpixel based upsampling increases accuracy or reduces inference time \cite{Imp_integration}.

In encoder-decoder models that employ superpixels, the superpixels are responsible for boundary alignment. We propose a regularization term that encourages the superpixels to be SLIC-superpixels \cite{SLIC}, that are based on pixel color and position, independent of the segmentation annotation. The proposed regularization term is simple and straightforward. We apply the regularization term to training an HCFCN-16 model and demonstrate that it significantly improves boundary alignment when trained on coarse annotations. The improved boundary alignment is not limited to coarse annotations and extends to fine annotations. Moreover, we demonstrate significant improvements to pixel accuracy in limited settings.
\section{Related work}

\subsection{Semantic segmentation}
Many applications, like autonomous driving, terrain segmentation and photo editing, rely on semantic segmentation. Semantic segmentation is the task of assigning a class to each pixel in an image. The Cityscapes \cite{Cityscapes}, PanNuke \cite{PanNuke}, SUIM \cite{SUIM} and COCO-Stuff \cite{Coco-Stuff} datasets are popular semantic segmentation benchmarks. These datasets provide fine segmentation annotations. With fine segmentation annotations, each pixel is annotated. Deep learning approaches \cite{FCN, FPN, UNet} have proven effective on the task of semantic segmentation with fine annotations. The Cityscapes dataset also provides coarse annotations. Coarsely annotating an image involves drawing simple polygons. With coarse annotations some pixels are left unlabeled. 

\subsection{Encoder-decoder models}
Many segmentation models employ an encoder-decoder structure. The encoder-decoder models \cite{FCN, FPN, SegNet, UNet, Deconvolution} use an encoder to extract high level features from the input image. Fully convolutional encoders, like ResNet \cite{ResNet}, are used to maintain the spatial structure. The resulting feature map has a lower spatial resolution than the input image. The decoder is tasked with upsampling this feature map to the desired segmentation output. Fully convolutional networks (FCNs) \cite{FCN} have set the ground work for encoder-decoder models. FCNs use convolutional layers and downsampling operations to extract high level features from an image. The encoder in FCN-16 reduces the spatial resolution to 1/16 the input resolution. A fully convolutional classifier brings the high dimensional features down to the desired number of classes. Finally, bilinear interpolation increases the spatial resolution to the final segmentation output. FCNs have proven effective for semantic segmentation, although they are prone to output imprecise segmentations. Bilinear interpolation is unable to recover much of the fine detail lost by the encoder. More recent approaches \cite{FPN, UNet, Deconvolution} employ learnable parameters in the decoder. These approaches are able to recover more detail, often with increased computational budget.

\subsection{Superpixels}
Superpixels are local clusters of pixels that share similar features. Superpixels are often used as a substitute for pixels as an efficient image representation. Simple Linear Iterative Clustering (SLIC) \cite{SLIC} is an effective algorithm for generating superpixels. SLIC employs K-means clustering to group nearby pixels into superpixels based on position and color features. Specifically, pixels are clustered in 5-D space defined by the x, y pixel coordinates and L, a, b values of the CIELAB color space. The superpixel size and compactness are both hyperparameters.

Several methods aim to integrate superpixels into neural networks. \cite{Bilateral_Inceptions} and \cite{SuperCNN} use pre-computed superpixels to increase model efficiency. SpixelFCN \cite{SpixelFCN} is a fully convolutional network that can be trained to output superpixels. In SpixelFCN a pixel does not belong to a single superpixel. Instead, a pixel belongs to nine local superpixels with various levels of association. SpixelFCN outputs an association map $\mathrm{Q} \in \mathbb{R}^{H \times W \times 9}$, which quantifies the pixel association to the surrounding superpixels. Two different choices of loss function are available to train SpixelFCN. The first loss function relies on segmentation annotations to learn superpixels. The second loss function imitates the SLIC objective and does not rely on segmentation annotations. \cite{SpixelFCN} proposes a downsampling/upsampling scheme to utilize the predicted superpixels in downstream tasks. Instead of feeding a high resolution image to a CNN. The image is first downsampled using the superpixels provided by SpixelFCN. The lower resolution image is used by the CNN in the downstream task. Finally the CNN output is upsampled using the same superpixels to recover the lost resolution. \cite{SpixelFCN} demonstrates the effectiveness of SpixelFCN by applying the scheme to PSMNet \cite{PSMNet} for the application of stereo matching.

\cite{Imp_integration} proposes the use of superpixels in encoder-decoder segmentation models by replacing the decoder with superpixel-based upsampling. This method significantly reduces inference time with a slight reduction in pixel accuracy when compared to state of the art models such as PSPNet \cite{PSPNet} and DeepLabv3 \cite{DeepLabv3}. Applying the proposed method to FCN, referred to as HCFCN, increases the pixel accuracy at the cost of inference time. \cref{hcfcn} depicts the architecture of HCFCN. We adopt the superpixels as proposed by \cite{Imp_integration} and compare to HCFCN-16, FCN-16, UNet and DeepLabv3+ as the baselines.

\subsection{Coarse annotations}
Supervised learning of convolutional neural networks is known to be data intensive, often requiring many annotated images to achieve satisfactory results. Semantic segmentation annotations in particular are very costly to collect. In a fine segmentation annotation, where each pixel is assigned a class, the object boundaries are difficult to annotate accurately \cite{LabelMe}. As a result, many data collectors opt for coarsely annotated images. Coarsely annotating an image involves drawing simple polygons. Each pixel within this polygon is assigned a common class. In \cref{label_sample}, a fine and a coarse annotation are shown. Fine annotations take over three times as much time as coarse annotations \cite{Label_quality}.

\cite{Label_quality} investigate the importance of annotation quality for semantic segmentation. Training on fine annotations yields superior results compared to training on coarse annotations. However, investing time in producing a large number of fine annotations is not the best data collection strategy. Ideally more time should be spent on producing coarse annotations than time spent on producing fine annotations. This strategy yields more annotations and results in superior pixel accuracy.

In this work, we aim to improve the boundary alignment of encoder-decoder models, that utilize superpixels as in \cite{Imp_integration}, when training annotations are coarse. The following section proposes a regularization term that encourages the superpixels to be SLIC-superpixels.

\section{Method}\label{sec:method}

\begin{figure*}[htbp]
\centerline{\includegraphics[width=0.96\textwidth]{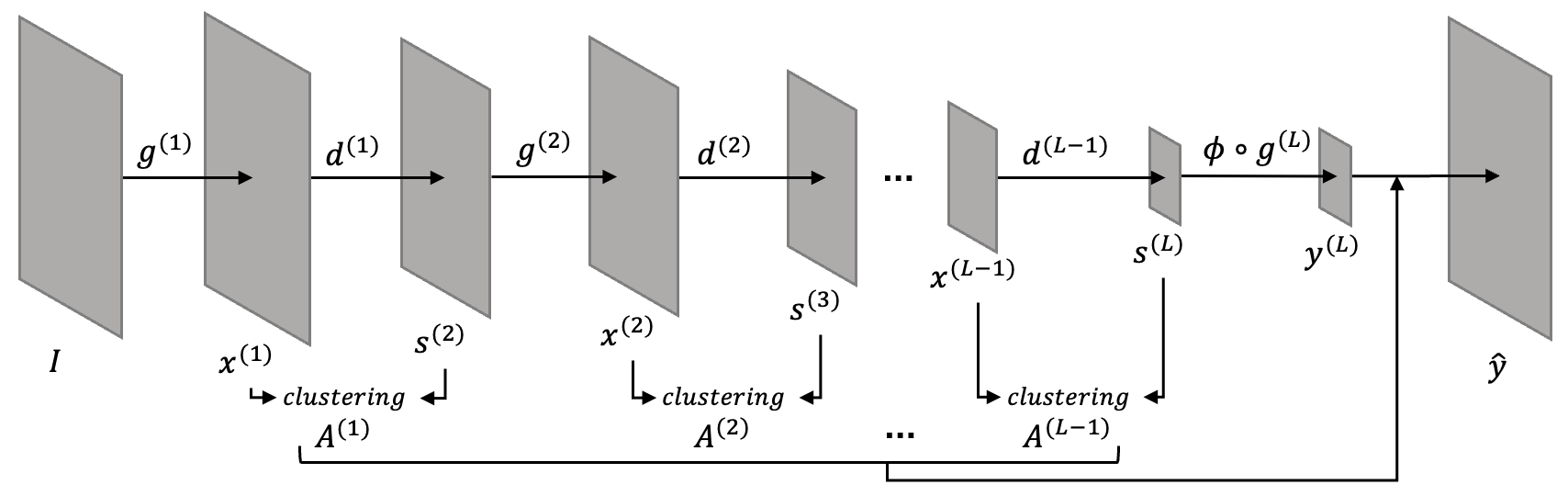}}
\caption{Architecture proposed by \cite{Imp_integration} with convolutional layers $g^{(l)}$ and downsampling layers $d^{(l)}$. Assignment matrix $A^{(l)}$ is obtained at the layer that downsamples $x^{(l)}$ to $s^{(l+1)}$. The low resolution output $y^{(L)}$ is upsampled by the assignment matrices $A$.}
\label{hcfcn}
\end{figure*}

\cite{Imp_integration} propose to replace the decoder in existing encoder-decoder model with superpixel based upsampling. They apply the idea to the FCN model and call it HCFCN, illustrated in \cref{hcfcn}. The encoder consists of convolution + ReLU layers $g^{(l)}$ and downsampling layers $d^{(l)}$, for $l \in {1, \ldots, L}$. The downsampling layer $d^{(l)}$ applied to the feature map $x^{(l)}$ results in the seed map $s^{(l+1)}$. The last convolution + ReLU layer $g^{(L)}$ is followed by a classifier $\phi$, resulting in the encoding $y^{(L)}$. The decoder upsamples the encoding $y^{(L)}$ into superpixels comprising the final segmentation $\hat{y}$:
\begin{equation}
\hat{y} = \prod_{l = L-1}^{1} A^{(l)} *  y^{(L)}.
\label{decoder}
\end{equation}
where the assignment matrices $A^{(l)}$ are computed during the forward pass through the encoder, which cluster the pixels in $x^{(l)}$ according to their similarity to the pixels in $s^{(l+1)}$. We will use $A = \{ A^{(l)} | l \in \{ 1, \ldots, L-1 \} \}$ to denote the set of all assignment matrices.

\subsection{SLIC regularization}
With coarse annotations, the pixels in the vicinity of class boundaries are often unlabeled. As such, the supervised loss provides little to no incentive for precise boundaries. We can introduce a bias by adding a regularization term to the loss function. The predicted class boundaries are determined by the superpixel boundaries. As such, the regularization term should be aimed at the superpixels. We seek a regularization term that introduces a bias for superpixels which is independent of the segmentation annotations. This bias should align the superpixel boundaries with the true class boundaries.

The SLIC algorithm clusters nearby pixels into superpixels based on CIELAB color and positional features. \cite{SpixelFCN} propose a loss function which imitates the SLIC objective. The loss function encourages the superpixels outputted by SpixelFCN to be SLIC-superpixels. The loss function proposed by \cite{SpixelFCN} does not apply to the superpixels in this work, because the superpixels in this work are embedded in the assignment matrices $A$. We propose a regularization term that encourages the superpixels given by the assignment matrices $A$ to be SLIC-superpixels.

\begin{figure} 
\centerline{\includegraphics[width=0.7\textwidth]{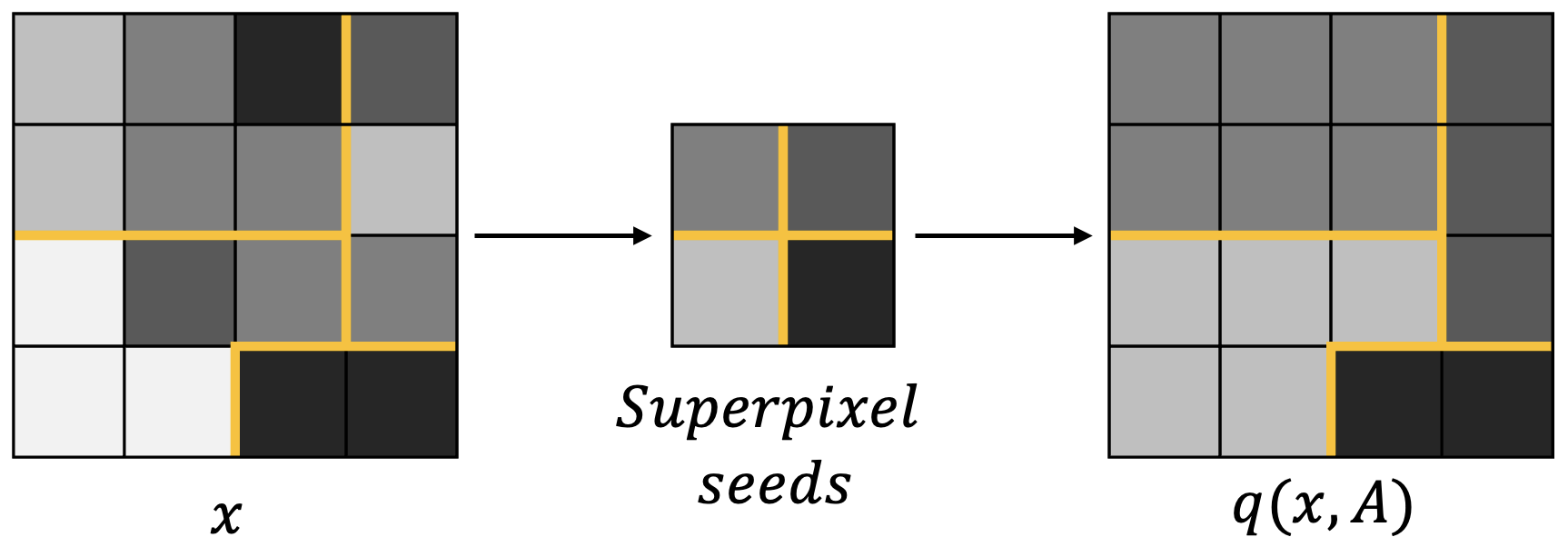}}
\caption{Illustration of the function $q(x, A)$ defined by equation \eqref{pooling}. The superpixel boundaries are highlighted in yellow. The feature map $x$ is downsampled using the the superpixels before upsampling with the same superpixels. For simplicity, each pixel is assigned to a single superpixel.}
\label{fig_pooling}
\end{figure}

Let us first define a general function $q(x, A)$ required by the regularization term:

\begin{equation}
q(x, A) = \prod_{l = L-1}^{1} A^{(l)} \prod_{l = 1}^{L-1} A^{(l)} * x
\label{pooling}
\end{equation}
where $x$ is a full resolution feature map, and $A$ is the set of all assignment matrices. It returns a full resolution feature map where each pixel contains the weighted average feature of its corresponding superpixels. First, the full resolution feature map $x$ is recursively downsampled using the assignment matrices $A$. The result is a low resolution feature map where each pixel is a superpixel seed. Each superpixel seed contains the weighted average feature computed over the pixels assigned to that superpixel. Then, this low resolution feature map is recursively upsampled with the same assignment matrices $A$, but in the opposite order. The result is a full resolution feature map. Each pixel in this feature map contains the weighted average feature of its corresponding superpixels. The function $q(x, A)$ is illustrated in \cref{fig_pooling}.

We use the function $q$ to construct the SLIC-superpixel regularization term as follows. Assuming the image $I$ is an RGB image, we first map $I$ to the CIELAB color space. The CIELAB color image is given by $f(I)$. It is then downsampled and upsampled with the assignment matrices $A$, using function $q$. Each pixel in the resulting feature map $q(f(I), A)$ holds the average CIELAB color features of its corresponding superpixels. For each pixel in the original image, we minimize the $\ell_2$ distance between its CIELAB color features and the average CIELAB color features of its corresponding superpixels. The resulting regularization term is given by equation \eqref{SLIC_loss}. 
\begin{equation}
L_{SLIC}(I, A) = \sum_{\textbf{p}} ||f(I)[\textbf{p}] - q(f(I), A)[\textbf{p}]||_2 \\
\label{SLIC_loss}
\end{equation}

The regularization term is added to the cross-entropy loss. The regularization strength is set by $\lambda$. The complete loss is given by equation \eqref{Total_loss}.

\begin{equation}
L_{total} = L_{cross-entropy} + \lambda * L_{SLIC}.
\label{Total_loss}
\end{equation}

Next to CIELAB color features, the SLIC algorithm uses spatial coordinates to encourage compact superpixels. Spatial coordinates could be included in the proposed regularization term to encourage compact superpixels. However, adding the spatial coordinates did not show any performance improvements in our experiments, and hence the spatial coordinates are not included in the proposed regularization term.

\section{Experimental results}\label{sec:results}
We train HCFCN-16 (i.e. $L = 4$ in \cref{sec:method}) with the proposed regularization term and compare against U-Net, DeepLabv3+, FCN-16 and HCFCN-16 without regularization.

\subsection{Experimental setup}
\noindent \textbf{Architecture.}\label{sec:architecture}
The proposed regularization term can be used in combination with any encoder-decoder segmentation model that employs superpixels as proposed by \cite{Imp_integration}. This work is limited to HCFCN-16. All models in this work employ a ResNet34 \cite{ResNet} encoder. The final layer, \textit{conv5\_x}, is excluded to ensure an output stride of 16. Following \cite{Imp_integration}, all downsampling operations are substituted with the modulated deformable convolution (DCNv2) \cite{Deconvolution} with a stride of two. As suggested in \cite{Imp_integration}, only assignment matrices $A^{(3)}$ and $A^{(4)}$ are learned, since there is little to no improvement in learning assignment matrices at the first two downsampling layers. Upsampling from $y^{3}$ to $\hat{y}$ is done with bilinear upsampling. The architectures of HCFCN-16 with and without regularization are identical.

Not all seeds in $s^{(l+1)}$ are considered for assignment, as this is not computationally applicable. Pixel $x^{(l)}_i$ is assigned to a subset of seeds in $s^{(l+1)}$ called the candidate seeds. In line with existing superpixel segmentation methods \cite{SLIC, Imp_integration, SpixelFCN} the candidate seeds are restricted to the nine surrounding seeds. \\

\noindent \textbf{Training.}
We employ an unbiased universal training method for all approaches. The results in \cref{boundary_align}, are obtained using the following training method. Cross-entropy is used as the supervised training loss. The regularization term $\lambda$ is set to $0.075$, $0.1$ and $0.05$ on the Cityscapes, PanNuke and SUIM datasets respectively\footnote{See the supplementary material.}. During training, the Cityscapes images are randomly cropped to a size of 512 by 512 pixels to fit into memory. At random, images are horizontally flipped. The optimizer of choice is Adam with a learning rate of $0.0005$. Training is done for 50 epochs on the Cityscapes and PanNuke datasets, and 75 epochs on the SUIM dataset. Training for 50 and 75 epochs allows the models to converge. The model with the highest validation accuracy is returned.

The results in \cref{results_sub,vary_coarsness} are obtained by training on the training set as described above before fine-tuning on the training and validation sets. Fine-tuning is done following algorithm 7.3 in \cite{finetune}. After training on the training set, the model is fine-tuned on the training and validation sets until the loss on the validation set matches the loss on training set before fine-tuning. As there is no guarantee that this criteria will be reached, fine-tuning is limited to 15 epochs on the Cityscapes and SUIM datasets. On the PanNuke dataset fine-tuning is limited to 25 epochs, due to the larger validation set size. \\

\begin{figure}
    \centering
    \subfloat[\centering Fine annotation]{{\includegraphics[width=0.28\textwidth]{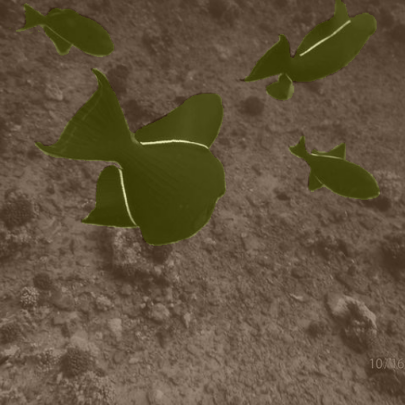}}}%
    \qquad
    \subfloat[\centering Eroded annotation]{{\includegraphics[width=0.28\textwidth]{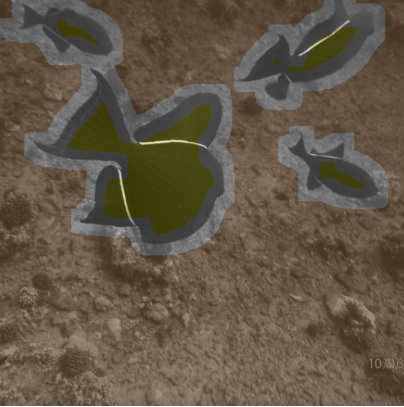}}}%
    \qquad
    \subfloat[\centering Coarse annotation]{{\includegraphics[width=0.28\textwidth]{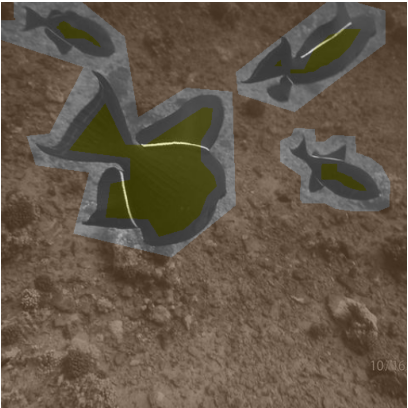}}}%
    \caption{Generating a coarse annotation from a fine annotation on the SUIM dataset. The fine annotation (a) is eroded by 12 pixels resulting in the eroded annotation (b). Finally the Douglas-pecker algorithm is used to approximate the eroded annotation with a polygon to obtain the coarse annotation (c).}%
    \label{make_coarse}%
\end{figure}

\begin{table}
\centering
\begin{tabular}{l|l|l|l|l}
Dataset    & Resolution & \# Train & \# Valid & \# Test \\ \hline
Cityscapes & 1024 $\times$ 512 & 2369 & 606 & 500 \\
SUIM & 512 $\times$ 512 & 1290 & 150 & 110 \\
PanNuke & 256 $\times$ 256 & 2620 & 2464 & 2655
\end{tabular}
\vspace{0.2cm}
\caption{Comparison of the Cityscapes, SUIM  and PanNuke dataset resolution and sizes of the training, validation and test sets.}
\label{datasets}
\end{table}

\noindent \textbf{Evaluation.}
All methods are evaluated with three semantic segmentation datasets: Cityscapes \cite{Cityscapes},  PanNuke \cite{PanNuke} and SUIM \cite{SUIM}. The Cityscapes dataset contains urban street scenes with detailed semantic segmentation annotations. The PanNuke dataset contains nuclei instance segmentations across different tissue types. The SUIM dataset contains semantic segmentation annotations for underwater images. A sample of each dataset can be seen in \cref{examples}. The proposed regularization term is expected to perform well on the SUIM dataset given its vibrant color palette.

The Cityscapes dataset offers fine and coarse annotations. All Cityscapes images are resized to 1024 by 512 pixels.  The resolution of the SUIM dataset is reduced by half to 512 by 512 pixels. The PanNuke and SUIM datasets do not provide coarse annotations. As such, coarse annotations are derived from the fine annotations, following \cite{Label_quality}. First, each class in the fine annotation is eroded by $6$ and $12$ pixels on the PanNuke and SUIM datasets respectively. Then, the Douglas-pecker algorithm is used to approximate the eroded annotation with a polygon. The resulting coarse annotations imitate the coarse annotations of the Cityscapes dataset. The process of generating a coarse annotation on the SUIM dataset is depicted in \cref{make_coarse}.

The SUIM datasets provides $110$ test samples. In this work, $150$ samples from the provided \textit{train\_val} set are reserved for validation. This leaves $1290$ samples in the training set. For the PanNuke dataset the first, second and third folds are used for the training, validation and test sets respectively. For the Cityscapes dataset, the provided validation set is used as the test set. The cities \textit{Aachen}, \textit{Monchengladbach}, \textit{Stuttgart} and \textit{Weimar} from the training set are chosen at random to be reserved for validation. \cref{datasets} shows the number of samples in each set for both datasets. All results in \cref{results_sub,vary_coarsness} are obtained by evaluating over the test set. In \cref{boundary_align}, all results are obtained by evaluating over the validation set.

All methods are evaluated using pixel accuracy (ACC) using: 
\[
ACC(\hat{y}, y) = \frac{\sum^{HW}_{i=1} \mathbb{1} \lbrace \hat{y}_i=y_i \rbrace}{\sum^{HW}_{i=1} \mathbb{1} \lbrace y_i \rbrace}
\label{pixel_acc}
\]
where $\hat{y}$ represents the segmentation output, $y$ represents the true segmentation, and $H$ and $W$ represent the height and width of $y$, respectively. Pixel accuracy is preferred over mIoU because of its lower sensitivity to classes that are underrepresented in an image. 

Boundary recall (BR) \cite{Boundary_Recall} is used to evaluate boundary alignment and is computed using:
\[
BR(\hat{y}, y) = \frac{n(b(\hat{y}, 0) \cap b(y, r))}{n(b(\hat{y}, 0) \cap b(y, r)) + n(b(y, 0) - b(\hat{y}, r))}
\]
where the function $b(y, r)$ returns the set of pixels within a $(2r + 1) \times (2r + 1)$ neighborhood of the boundary pixels in $y$. Following \cite{superpixel_eval}, $r$ is set to 0.0025 times the image diagonal. Boundary recall is preferred over Mean Squared Error (MSE), because of the significant computational complexity of MSE.



All results are obtained by averaging over five training and evaluation runs. When a difference between methods is described as significant, it is supported by a one-sided Mann–Whitney U Test with $\alpha = 0.05$.

\subsection{Results}\label{results_sub}
\begin{figure}
\captionsetup[subfigure]{labelformat=empty}
\centering
\subfloat{\includegraphics[width=.16\linewidth]{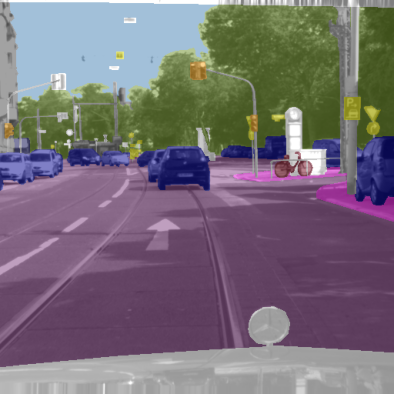}}
    \hfill
\subfloat{\includegraphics[width=.16\linewidth]{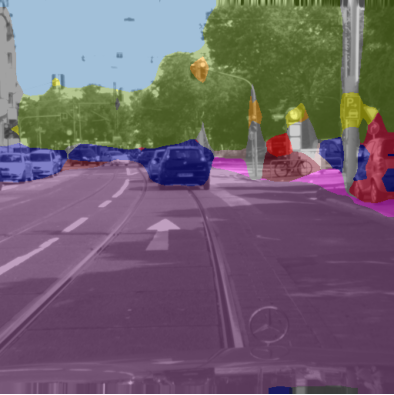}}
    \hfill
\subfloat{\includegraphics[width=.16\linewidth]{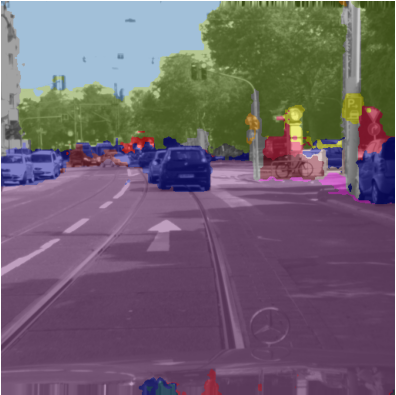}}
    \hfill
\subfloat{\includegraphics[width=.16\linewidth]{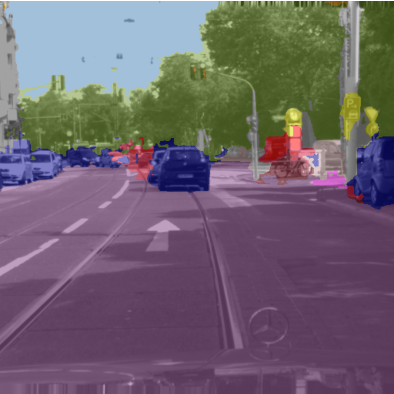}}
    \hfill
\subfloat{\includegraphics[width=.16\linewidth]{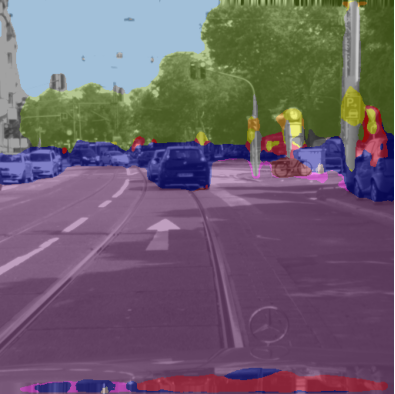}}
    \hfill
\subfloat{\includegraphics[width=.16\linewidth]{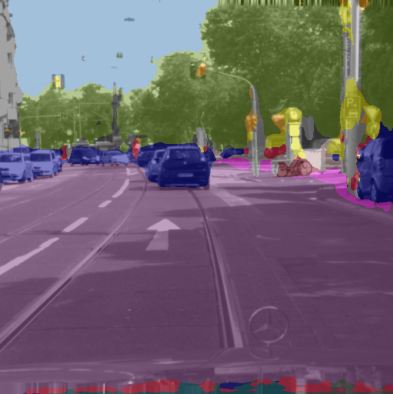}}

\subfloat{\includegraphics[width=.16\linewidth]{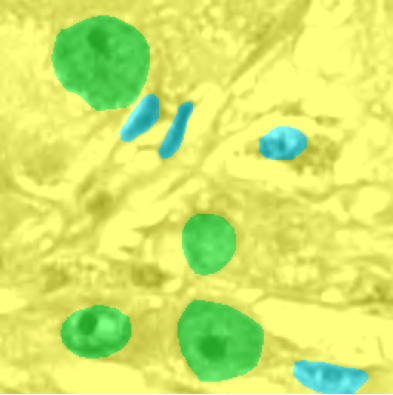}}
    \hfill
\subfloat{\includegraphics[width=.16\linewidth]{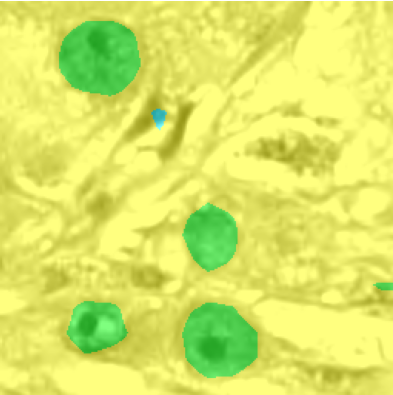}}
    \hfill
\subfloat{\includegraphics[width=.16\linewidth]{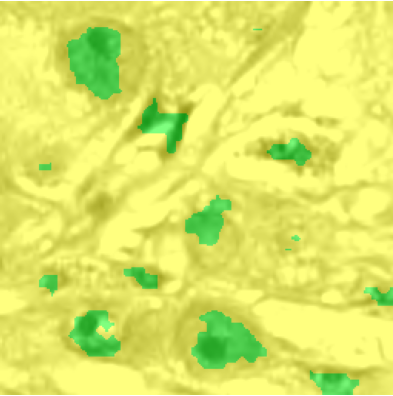}}
    \hfill
\subfloat{\includegraphics[width=.16\linewidth]{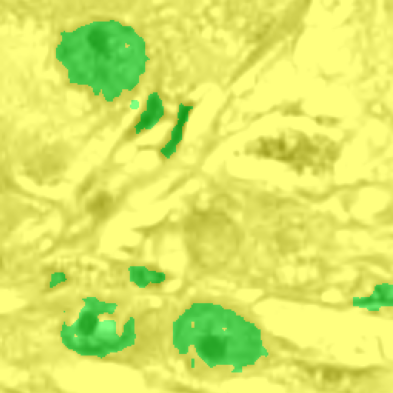}}
    \hfill
\subfloat{\includegraphics[width=.16\linewidth]{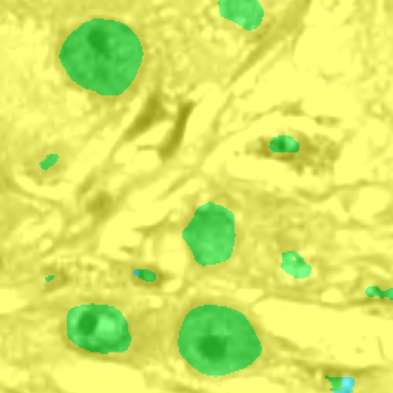}}
    \hfill
\subfloat{\includegraphics[width=.16\linewidth]{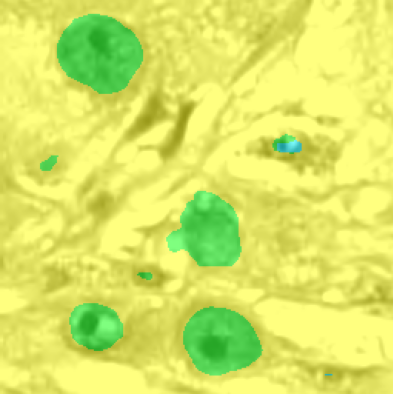}}

\subfloat[\centering Ground truth]{\includegraphics[width=.16\linewidth]{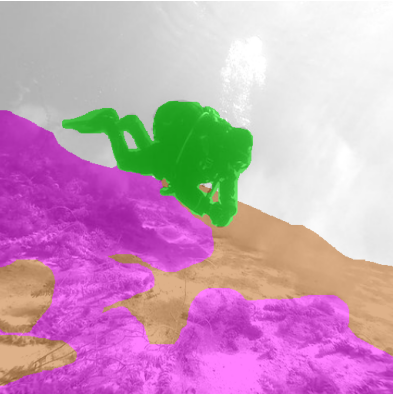}}
    \hfill
\subfloat[\centering FCN-16]{\includegraphics[width=.16\linewidth]{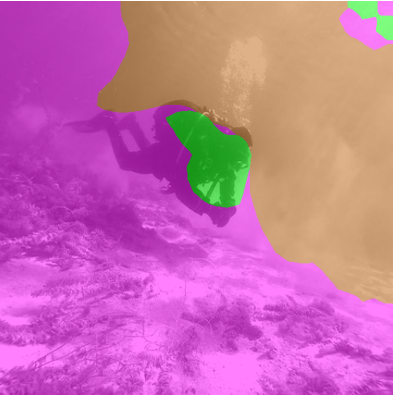}}
    \hfill
\subfloat[\centering HCFCN-16]{\includegraphics[width=.16\linewidth]{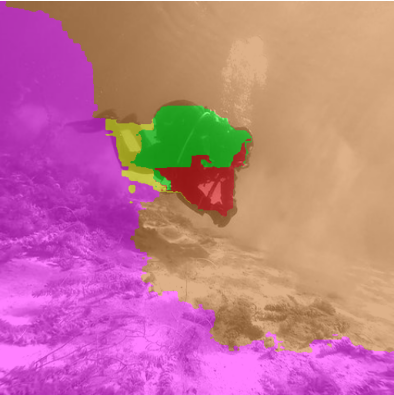}}
    \hfill
\subfloat[\centering Regularized HCFCN-16]{\includegraphics[width=.16\linewidth]{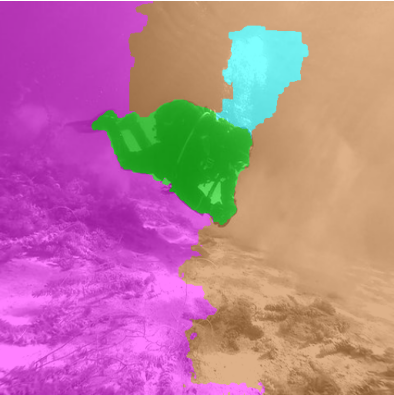}}
    \hfill 
\subfloat[\centering U-Net]{\includegraphics[width=.16\linewidth]{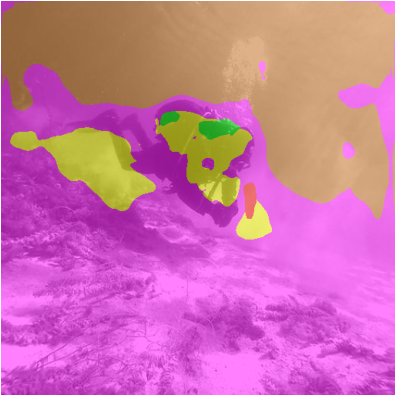}}
    \hfill
\subfloat[\centering DeepLabv3+]{\includegraphics[width=.16\linewidth]{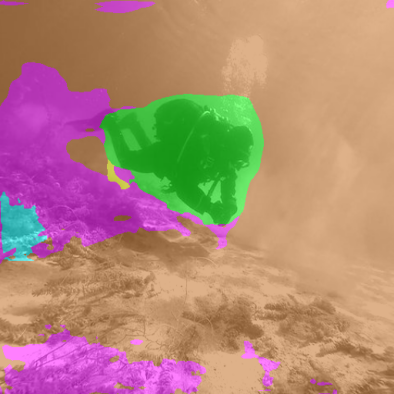}}
\caption{Example results on the Cityscapes (top), PanNuke (middle) the SUIM (bottom) test sets. All methods are trained on coarse annotations and fine-tuned on the training and validation set. The Cityscapes images are cropped to match the PanNuke and SUIM images. Examples are chosen to highlight the impact of regularization.}
\label{examples}
\end{figure}

In \cref{results_city,results_pannuke,results_suim} all methods are evaluated on the Cityscapes, PanNuke and SUIM test sets respectively. On the SUIM dataset the regularized HCFCN-16 achieves significantly higher accuracy and boundary recall than all other methods when trained on fine or coarse annotations. On the PanNuke and Cityscapes datasets the improvements are limited to boundary recall when trained on coarse annotations. In \cref{examples} an example of each method for each dataset is shown. FCN-16 and HCFCN-16 both fail to provide precise segmentations on the SUIM dataset. On the Cityscapes and PanNuke sample the impact of regularization is limited. This is in line with the results seen in \cref{results_city,results_pannuke,results_suim}.

\begin{table}[p]
\centering
\begin{tabular}{p{16mm} | p{20mm} |l p{22mm}| l p{22mm}} & & \multicolumn{2}{l}{Fine training annotations} & \multicolumn{2}{l}{Coarse training annotations}  \\
\hline
Data set & Method & Pixel acc. & BR & Pixel acc. & BR \\
\hline
Cityscapes & U-Net & 87.4 ± 0.48 & \textbf{65.1 ± 0.48} & 84.8 ± 0.18 & 46.8 ± 0.99 \\
 & DeepLabv3+ & 88.1 ± 0.42 & 64.6 ± 0.77 & 84.9 ± 0.41 & 46.2 ± 1.41 \\
 & FCN-16 & 87.9 ± 0.3 & 48.0 ± 1.09 & 84.5 ± 0.36 & 34.0 ± 0.44 \\
 & HCFCN-16 & \textbf{88.2 ± 0.28} & 61.9 ± 0.72 & 84.8 ± 0.19 & 47.1 ± 1.89 \\
 & Regularized HCFCN-16 & 88.2 ± 0.20 & 63.6 ± 0.92 & \textbf{85.3 ± 0.18} & \textbf{54.6 ± 0.90} \\
\hline
PanNuke & U-Net & \textbf{92.4 ± 0.10} & \textbf{51.2 ± 1.09} & 88.1 ± 0.19 & 6.3 ± 0.73 \\
 & DeepLabv3+ & 92.4 ± 0.03 & 47.7 ± 0.74 & 88.2 ± 0.22 & 6.3 ± 0.65 \\
 & FCN-16 & 90.2 ± 0.08 & 28.7 ± 0.26 & \textbf{88.6 ± 0.06} & 12.2 ± 0.57 \\
 & HCFCN-16 & 89.3 ± 1.36 & 34.7 ± 5.41 & 85.3 ± 1.23 & 6.2 ± 0.84 \\
 & Regularized \newline HCFCN-16 & 89.3 ± 0.72 & 35.1 ± 3.03 & 86.9 ± 0.51 & \textbf{14.4 ± 0.94} \\
\hline
SUIM & U-Net & 68.8 ± 1.56 & 13.1 ± 1.15 & 65.1 ± 1.38 & 7.4 ± 0.68 \\
 & DeepLabv3+ & 69.7 ± 1.19 & 13.2 ± 0.76 & 64.1 ± 1.63 & 7.4 ± 0.99 \\
 & FCN-16 & 71.1 ± 1.00 & 6.6 ± 0.43 & 68.4 ± 0.78 & 4.8 ± 0.24 \\
 & HCFCN-16 & 70.8 ± 1.08 & 14.4 ± 1.05 & 63.5 ± 4.67 & 7.8 ± 2.09 \\
 & Regularized \newline HCFCN-16 & \textbf{72.9 ± 0.34} & \textbf{18.7 ± 0.70} & \textbf{70.4 ± 0.66} & \textbf{12.5 ± 0.63} \\
\end{tabular}
\vspace{0.3cm}
\caption{Average pixel accuracy and boundary recall on the Cityscapes, PanNuke and SUIM test set, when trained on fine and coarse annotations. All methods are fine-tuned on the training and validation sets. Results are obtained by averaging over five training and evaluation runs. }
\label{results_city}
\label{results_pannuke}
\label{results_suim}
\label{tab:results}
\end{table}

\subsection{Varying level of coarseness}\label{vary_coarsness}
\begin{figure}[htbp]
    \centering
    \begin{subfigure}[b]{\textwidth}
        \noindent
        \begin{minipage}{0.5\textwidth}
            \includegraphics[width=\textwidth]{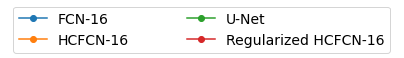}
            \includegraphics[width=\textwidth]{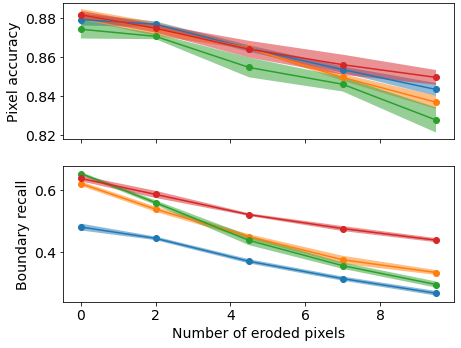}
        \end{minipage}
        \hspace{0.02\textwidth}
        \begin{minipage}{0.5\textwidth}
            \vspace{0.9cm}
            \subfloat{\includegraphics[width=.33\textwidth]{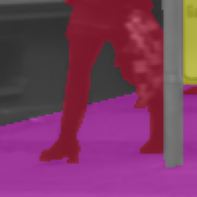}}
                \hfill
            \subfloat{\includegraphics[width=.33\textwidth]{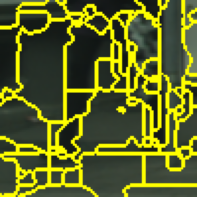}}
                \hfill
            \subfloat{\includegraphics[width=.33\textwidth]{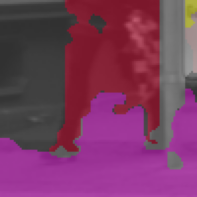}}
            \captionsetup{labelformat=empty}
            \subfloat[\centering Label]{\includegraphics[width=.33\textwidth]{Figures/city_bound_label.png}}
                \hfill
            \subfloat[\centering Superpixels]{\includegraphics[width=.33\textwidth]{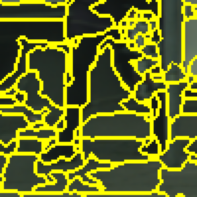}}
                \hfill
            \subfloat[\centering Output]{\includegraphics[width=.33\textwidth]{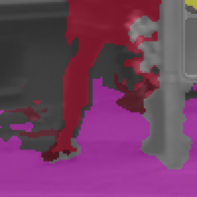}}
        \end{minipage}
        \vspace{-0.2cm}
        \setcounter{subfigure}{0}
        \caption{Cityscapes}
        \label{city_vary_coarse}
    \end{subfigure}
    \hfill
    
    \begin{subfigure}[b]{\textwidth}  
        \noindent
        \begin{minipage}{0.5\textwidth}
            \includegraphics[width=\textwidth]{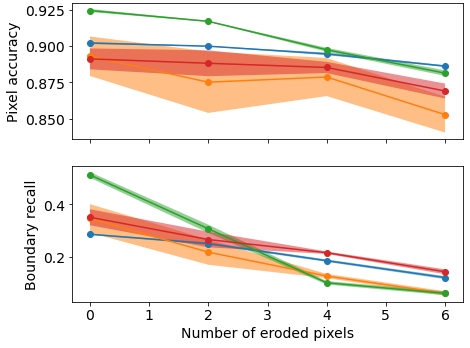}
        \end{minipage}
        \hspace{0.02\textwidth}
        \begin{minipage}{0.5\textwidth}
            \subfloat{\includegraphics[width=.33\linewidth]{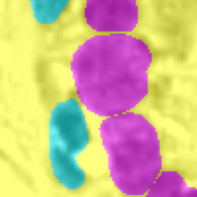}}
                \hfill
            \subfloat{\includegraphics[width=.33\linewidth]{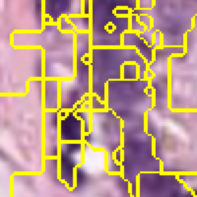}}
                \hfill
            \subfloat{\includegraphics[width=.33\linewidth]{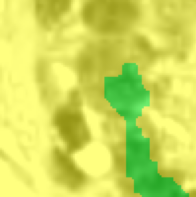}}
            \captionsetup{labelformat=empty}
            \subfloat[\centering Label]{\includegraphics[width=.33\linewidth]{Figures/pannuke_bound_label.png}}
                \hfill
            \subfloat[\centering Superpixels]{\includegraphics[width=.33\linewidth]{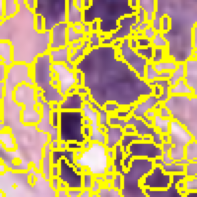}}
                \hfill
            \subfloat[\centering Output]{\includegraphics[width=.33\linewidth]{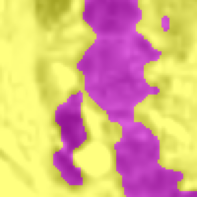}}
        \end{minipage}
        \vspace{-0.2cm}
        \setcounter{subfigure}{1}
        \caption{PanNuke}
        \label{pannuke_vary_coarse}
    \end{subfigure}
    \hfill
    
    \begin{subfigure}[b]{\textwidth}  
        \noindent
        \begin{minipage}{0.5\textwidth}
            \includegraphics[width=\textwidth]{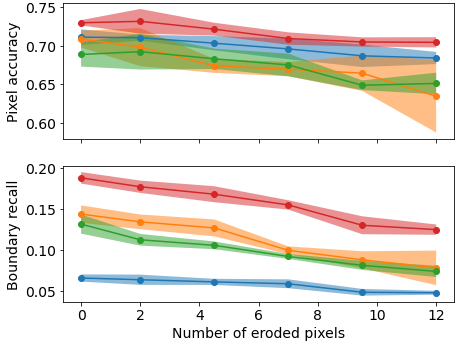}
        \end{minipage}
        \hspace{0.02\textwidth}
        \begin{minipage}{0.5\textwidth}
            \subfloat{\includegraphics[width=.33\linewidth]{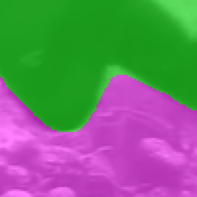}}
                \hfill
            \subfloat{\includegraphics[width=.33\linewidth]{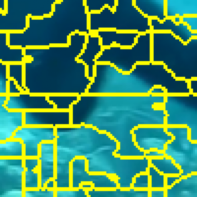}}
                \hfill
            \subfloat{\includegraphics[width=.33\linewidth]{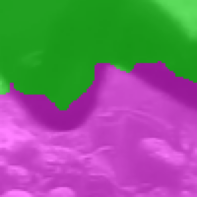}}
            \captionsetup{labelformat=empty}
            \subfloat[\centering Label]{\includegraphics[width=.33\linewidth]{Figures/suim_bound_label.png}}
                \hfill
            \subfloat[\centering Superpixels]{\includegraphics[width=.33\linewidth]{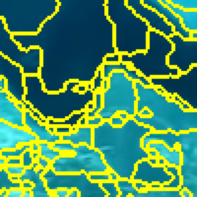}}
                \hfill
            \subfloat[\centering Output]{\includegraphics[width=.33\linewidth]{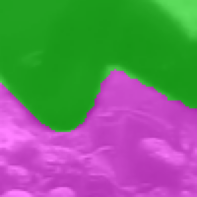}}
        \end{minipage}
        \vspace{-0.2cm}
        \setcounter{subfigure}{2}
        \caption{SUIM}
        \label{suim_vary_coarse}
    \end{subfigure}
    
    \caption{\label{boundary}Left: pixel accuracy and boundary recall on the Cityscapes, PanNuke and SUIM test sets when trained on coarse annotations with increasing number of eroded pixels. Right: examples of boundary alignment on the validation sets, with HCFCN-16 in the top and Regularized HCFCN-16 in the bottom row.}
\end{figure}

We investigate the impact of intermediate levels of coarseness. The intermediate levels of coarseness are obtained by varying the number of eroded pixels. In \cref{city_vary_coarse,pannuke_vary_coarse,suim_vary_coarse} we can see the impact on pixel accuracy and boundary recall when trained on increasingly coarse annotations. All models are trained on the coarse training set and fine-tuned on the coarse training and validation sets.

\subsection{Boundary alignment}\label{boundary_align}
The regularization term aims to improve boundary alignment. The improved boundary recall suggests the regularization term is effective in doing so. We investigate the role of the superpixels in improving the boundary alignment. In \cref{city_vary_coarse} a region of a Cityscapes sample is highlighted. The annotation on the left shows the intricate class boundaries of this sample. The predicted boundaries of HCFCN-16 do not align with the ground truth boundaries exactly. Regularization arguably improves boundary alignment in this example. The superpixels of both the HCFCN-16 and the regularized HCFCN-16 show reasonable boundary alignment. The imperfect boundary alignment is a result of misclassified superpixels and not misaligned superpixels. 

In \cref{pannuke_vary_coarse} a region of a PanNuke sample is shown. The regularization term impacts the superpixels of HCFCN-16. The superpixel boundaries adhere to the color boundaries present in the image. 

In \cref{suim_vary_coarse} a region of a SUIM sample is shown. The ground truth class boundary is simple compared to the Cityscapes sample. Still HCFCN-16 fails to align the predicted boundary with the ground truth boundary. The superpixels, plotted in the middle, explain the imprecise segmentation. The superpixels in HCFCN-16 do not align with the ground truth boundary. The superpixels of the regularized HCFCN-16 do align with the ground truth boundary. As a result the predicted boundary in the the regularized HCFCN-16 output aligns with the ground truth boundary.

\section{Discussion}
Regularizing the superpixels in HCFCN-16 significantly improves boundary alignment on all three data sets with respect to the baseline methods, as shown by an improvement in boundary recall in \cref{tab:results}. Boundary recall improves up to 60.3\% on the SUIM dataset w.r.t. the next best method. \cref{boundary}c illustrates the improved boundary alignment of the regularized HCFCN-16 on the SUIM dataset, where without regularization, HCFCN-16 suffers from poor boundary alignment.

The impact of regularization varies between the datasets included in this work. The impact of regularization is larger on the SUIM dataset than on the Cityscapes and PanNuke datasets, where the baseline methods already achieve relatively high boundary recall without regularization. Results in in \cref{results_city,results_pannuke,results_suim} suggest that regularizing the superpixels is most effective on difficult data sets, when other models suffer from poor boundary alignment.


On the Cityscapes and the PanNuke datasets, when trained on fine annotations, U-Net and DeepLabv3+ achieve significantly higher boundary recall than the regularized HCFCN-16. The additional learnable parameters in the U-Net and DeepLabv3+ decoders allow for an improved boundary recall. When training annotations are coarse, however, the pixel accuracy and boundary recall of U-Net and DeepLabv3+ drop significantly.

The impact on training time is minimal. The regularized HCFCN-16 takes 3.8\% longer to complete a single epoch over the Cityscapes training set than HCFCN-16 without regularization using a NVIDIA Tesla T4 GPU.

\section{Conclusion}
Boundary alignment in semantic segmentation is important for various applications, including semantic segmentation of medical images, quality control in high-tech manufacturing, or photo/video editing. In this work we propose a regularization term that improves boundary alignment when training on coarse annotations, for cases where obtaining fine annotations is expensive or difficult. The regularization term encourages SLIC-superpixels in encoder-decoder models that employ a superpixel based decoder. The proposed method was evaluated on the Cityscapes, PanNuke and SUIM datasets. Boundary recall improves significantly on all datasets when trained on coarse annotations, with an improvement of 60.3\% on the SUIM dataset w.r.t. the next best method. Pixel accuracy improves as well when trained coarse annotations. With coarse annotations being easier and cheaper to obtain, the proposed method can lower the cost of developing semantic segmentation models.


\section{Ethical statement}
The data used in this work comes from standard data sets that are publicly available. The PanNuke and SUIM contain no personal or
sensitive information. The Cityscapes data set contains some personal information (such as car license plates), however, this information is not relevant nor used in any way by the proposed method. No personal information is inferred. This work proposes a general method for semantic segmentation with coarse annotations and does not target nor was it inspired by any policing or military applications.

\bibliographystyle{splncs04}
\bibliography{references}
\newpage
\section{Supplementary Material}

\subsection{Hyperparameter study}
\begin{figure}
    \centering
    \subfloat[\centering Accuracy]{{\includegraphics[width=0.47\textwidth]{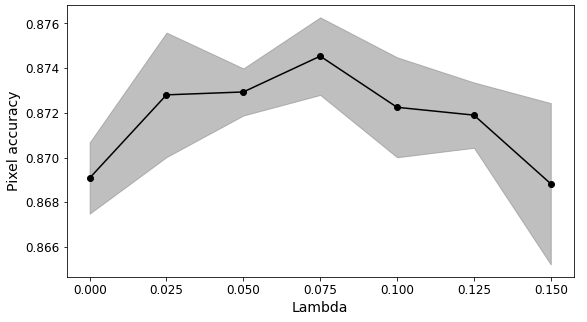}}}%
    \qquad
    \subfloat[\centering Boundary recall]{{\includegraphics[width=0.47\textwidth]{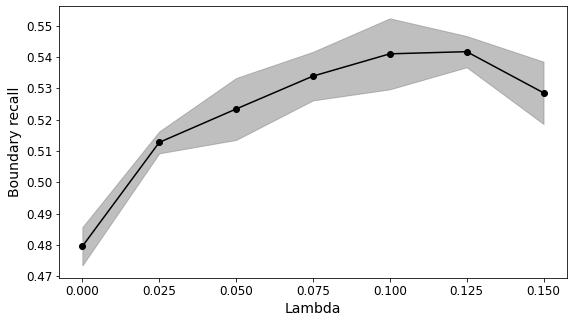}}}%
    \caption{Average accuracy (a) and boundary recall (b) on the Cityscapes validation set for various levels of $\lambda$. HCFCN-16 is trained on Cityscapes coarse training annotations.}%
    \label{lambda_hyper}%
\end{figure}

\noindent \textbf{Hyperparameter $\lambda$.} We study the impact of $\lambda$ when trained on Cityscapes coarse annotations. With $\lambda$ the regularization strength can be adjusted. In \cref{lambda_hyper} (a) we can see that pixel accuracy starts to degrade when $\lambda$ surpasses $0.075$. In \cref{lambda_hyper} (b) we can see boundary recall still benefiting from an increased $\lambda$. The maximum boundary recall is achieved at a $\lambda$ of 0.125 on the Cityscapes dataset. For the Cityscapes dataset, $\lambda$ is set to $0.075$ to maximise pixel accuracy. A similar relation is concluded on the PanNuke and SUIM datasets. On the PanNuke and SUIM datasets the maximum pixel accuracy is achieved when $\lambda$ is set to $0.1$ and $0.05$ respectively. \\

\subsection{Including spatial coordinates}\label{sec:spatial_coordinates}
The SLIC algorithm uses spatial coordinates to encourage compact superpixels. Similarly spatial coordinates can be included in the proposed regularization term to encourage compact superpixels. Let $\textbf{p} = (i, j)$ be a pixel in the original image (with $(i, j)$ being its image coordinates) and $\textbf{P}$ be the map of all pixel coordinates. Using function $q$, we obtain the feature map $q(\textbf{P}, A)$, where each pixel in the feature map holds the \emph{average position} of its corresponding superpixels. To encourage compact superpixels, the $\ell_2$ distance between the pixel position and the average position of its corresponding superpixels is minimized. This term can be added to the regularization term with hyperparameter \textit{m} to determine the compactness. The resulting regularization term is given by \cref{SLIC_loss_with_m}.

\begin{equation}
\begin{aligned}
L_{SLIC}(I, A) = & \sum_{\textbf{p}} ||f(I)[\textbf{p}] - q(f(I), A)[\textbf{p}]||_2 \\
& + m * ||\textbf{p} - q(\textbf{P}, A)[\textbf{p}]||_2.
\end{aligned}
\label{SLIC_loss_with_m}
\end{equation}

\begin{figure}
    \centering
    \subfloat[\centering Accuracy]{{\includegraphics[width=0.47\textwidth]{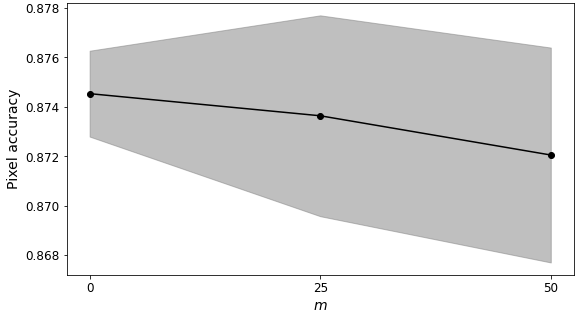}}}%
    \qquad
    \subfloat[\centering Boundary recall]{{\includegraphics[width=0.47\textwidth]{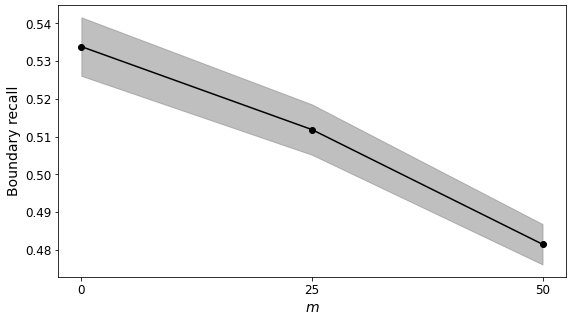}}}%
    \caption{Average accuracy (a) and boundary recall (b) on the Cityscapes validation set for various levels of \textit{m}. HCFCN-16 is trained on Cityscapes coarse training annotations.}%
    \label{m_hyper}%
\end{figure}

Increasing hyperparameter \textit{m}, encourages superpixels to be more compact. Low values for \textit{m} can lead to scattered superpixels. In \cref{m_hyper} the pixel accuracy and boundary recall for various levels of \textit{m} are plotted. The regularization term is most effective when \textit{m} is set to zero. This relation holds for the PanNuke and SUIM datasets as well. Setting \textit{m} to zero does not lead to scattered superpixels. This could be explained by the supervised loss encouraging compact superpixels, making the inclusion of spatial coordinates in the regularization term redundant. 

\end{document}